\documentclass[letterpaper, 10 pt, conference]{ieeeconf}  % Comment this line out if you need a4paper

\IEEEoverridecommandlockouts                              % This command is only needed if 
\usepackage{graphics} % for pdf, bitmapped graphics files
\usepackage{epsfig} % for postscript graphics files
\usepackage{mathptmx} % assumes new font selection scheme installed
\usepackage{times} % assumes new font selection scheme installed
\usepackage{amsmath} % assumes amsmath package installed
\usepackage{amssymb}  % assumes amsmath package installed

\usepackage[switch]{lineno}
\usepackage{booktabs}
\usepackage{threeparttable}
\usepackage{romannum}
\usepackage{soul}

\usepackage{amsthm,amsmath,amssymb}
\usepackage{mathrsfs}

\usepackage{color, xcolor}
\usepackage{here}
\usepackage{graphicx}
\usepackage{subfigure}
\usepackage{cite}
\usepackage{bm}
\usepackage{float}
\usepackage{mathtools}
\usepackage{amsmath,amsfonts}
\usepackage{algorithm}
\newcommand{\nonl}{\renewcommand{\nl}{\let\nl\oldnl}}% 

\title{\LARGE \bf
Adaptive Model Predictive Control with Data-driven Error Model for Quadrupedal Locomotion
}

\author{ Xuanqi Zeng$^{1}$, Hongbo Zhang$^{1}$, Linzhu Yue$^{1}$, Zhitao Song$^{1}$, Lingwei Zhang$^{2}$ and Yun-Hui Liu$^{1}$ % <-this % stops a space
\thanks{$^{1}$ X. Q. Zeng, H. B. Zhang, L. Z. Yue, Z. T. Song and Y.-H. Liu are with the Department of Mechanical and Automation Engineering, The Chinese University of Hong Kong.
        {\tt\small xqzeng@mae.cuhk.edu.hk}}%
\thanks{$^{2}$ L. W. Zhang is with the Hong Kong Centre for Logistics Robotics.}
\thanks{* Corresponding author: Y.-H. Liu  
        {\tt\small yhliu@cuhk.edu.hk}}
\thanks{The InnoHK Clusters support this work via the Hong Kong Centre of Logistics Robotics.}% <-this % stops a space
}

\begin{document}
\maketitle
\thispagestyle{empty}
\pagestyle{empty}

%%Due to the similar structure with its animal counterpart, legged robot is expected as a possible solution to replace human to do some operations in today complex human-domain scenarios.
%%Thus, it is essential to develop a robust controller for legged robot which can let robot to have good performance in the real world.
%%%%%%%%%%%%%%%%%%%%%%%%%%%%%%%%%%%%%%%%%%%%%%%%%%%%%%%%%%%%%%%%%%%%%%%%%%%%%%%%
\begin{abstract}
Model Predictive Control (MPC) relies heavily on the robot model for its control law. However, a gap always exists between the reduced-order control model with uncertainties and the real robot, which degrades its performance. To address this issue, we propose the controller of integrating a data-driven error model into traditional MPC for quadruped robots. Our approach leverages real-world data from sensors to compensate for defects in the control model. Specifically, we employ the Autoregressive Moving Average Vector (ARMAV) model to construct the state error model of the quadruped robot using data. The predicted state errors are then used to adjust the predicted future robot states generated by MPC. By such an approach, our proposed controller can provide more accurate inputs to the system, enabling it to achieve desired states even in the presence of model parameter inaccuracies or disturbances. The proposed controller exhibits the capability to partially eliminate the disparity between the model and the real-world robot, thereby enhancing the locomotion performance of quadruped robots. We validate our proposed method through simulations and real-world experimental trials on a large-size quadruped robot that involves carrying a 20 kg un-modeled payload (84\% of body weight).

\end{abstract}

%%%%%%%%%%%%%%%%%%%%%%%%%%%%%%%%%%%%%%%%%%%%%%%%%%%%%%%%%%%%%%%%%%%%%%%%%%%%%%%%

\section{INTRODUCTION}
%Quadruped robots, due to their structural similarity with animals, offer a promising solution for performing complex tasks in various human-domain scenarios. However, compared to traditional wheeled mobile robots, controlling legged robots presents a more intricate challenge. The control methods for quadruped robots can be broadly categorized into two groups: model-based control and learning-based control.
Quadruped robots’ promising application potential lets them become a research hotspot in the robotic field, and numerous research outcomes about it have sprung up in recent years \cite{Di_01} -\cite{our_02}. For the model-based control of the quadruped robot, the controller is designed based on the robot's dynamic model, which necessitates accurate model parameters. However, on the one hand, these parameters come from computer-aided design (CAD) software or simplified models due to computation costs in most real-world applications. On the other hand, certain parts of the system are challenging to include in physical models: 1) Friction in mechanical transmission; 2) Nonlinear output from the actuator; 3) Mechanical deformation under impact; 4) Unknown disturbances from the environment (such as un-modeled payload). These facts can result in inaccuracies in the control model. Although the controller's robustness may enable it to operate despite the defect model, these constraints can negatively impact the locomotion performance of robots in real-world scenarios.

The objective of this paper is to employ real-world data to address the issue of model uncertainty in a model-based controller. The measurements extracted from onboard sensors can be considered as ground truth to some extent, and they reveal the defective part of the model. Thus, we propose to construct an error model by data-driven method in the view of time-series analysis, then use it to anticipate future errors and correct them proactively. In such ways, the real-world quadrupedal locomotion can be improved even under un-modeled payloads.

\begin{figure}[t]
  \centering
\vspace{-0.2cm}
 \subfigure[Baseline controller, 10 kg payload]{
    \label{}
    \includegraphics[width=0.45\textwidth]{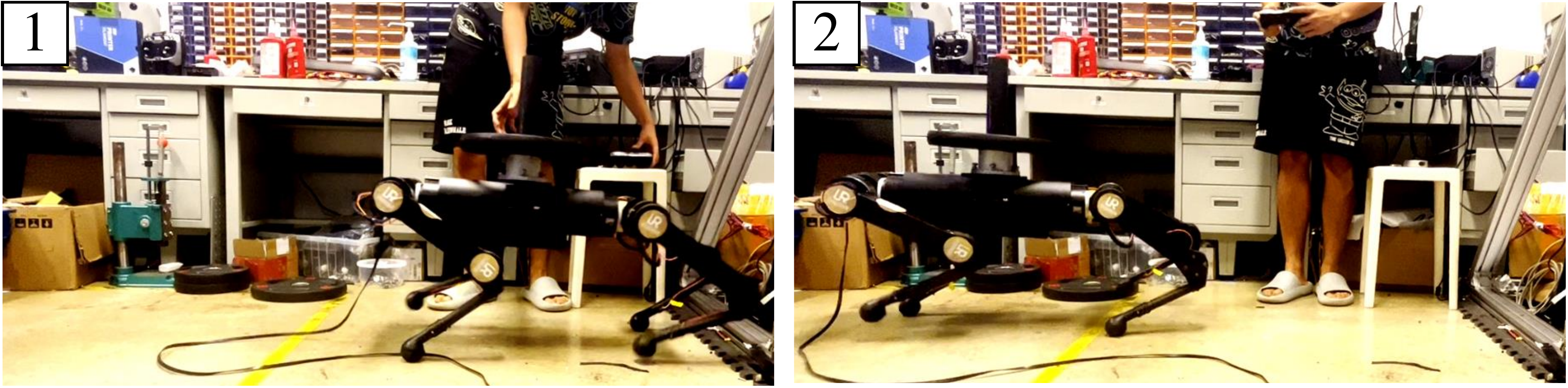}}
  \subfigure[Proposed controller, 20 kg payload]{
    \label{} %% label for first subfigure
    \includegraphics[width=0.45\textwidth]{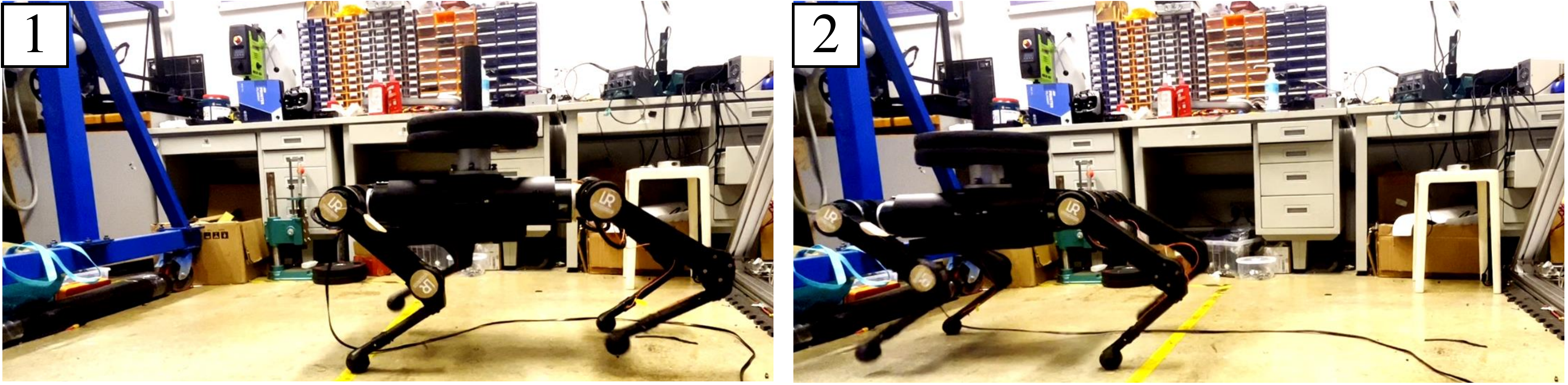}}
  % \subfigure[Baseline(fall down)]{
  %   \label{sim_baseline_fall} %% label for first subfigure
  %   \includegraphics[width=0.15\textwidth]{baseline_payload8_falldown.png}}
\vspace{-3mm}
  \caption{The forward trotting with un-modeled payload experiment. (a) Baseline: the mean of body height is 0.367 m (desired value is 0.4 m) and its STD (Standard Deviation) is 0.0035 m; (b) Proposed: the mean of body height is 0.397 m and its STD is 0.0011 m.}
  \label{forward_trotting} %% label for entire figure
\vspace{-0.3cm}
\end{figure}

\iffalse
\begin{center}
\vspace{-0.6cm}
\begin{figure}[t]
\centering
\includegraphics[width=3.3in]{Fig_1.pdf}
\caption{The forward trotting with un-modeled payload experiment. The upside is the snapshot of the robot carrying a 10 kg payload with the baseline controller, where the mean body height is 0.367 m (desired value is 0.4 m) and its STD (Standard Deviation) is 0.0035 m. The snapshot of the robot with the proposed controller  carrying a 20 kg payload is shown below part, where the body height is 0.397 m and its STD is 0.0011 m.}\label{forward_trotting}
\end{figure}
\vspace{-0.6cm}
\end{center}
\fi

\subsection{Related Work} \label{related_work}
\paragraph{Model Predictive Control}
Model-based control methods, particularly Model Predictive Control (MPC), have garnered significant attention in recent research on quadruped robot control. It has demonstrated impressive performance in controlling quadruped robots. For instance, the MIT Cheetah 3 and Mini-Cheetah, utilizing convex MPC, achieves robust locomotion in different gaits (e.g., trot, bound, gallop) \cite{Di_01}, \cite{Kim_02}.  However, it is worth noting that their works are based on the simplifying assumption of a Single Rigid Body (SRB), due to the trade-off between model accuracy and computational cost in MPC. To address this, there are some researches about adopting a full-dynamic system model that implements Nonlinear Model Predictive Control (NMPC) in quadruped robots to improve control accuracy  \cite{Neunert_03}, \cite{Meduri_04}.

\paragraph{Adaptive Control}
This approach tackles the time-varying model by adjusting the model or control parameters online. To illustrate, the IIT group employs an offline and online identification algorithm on HyQ using a recursive approach \cite{Tournois_11}. Minniti et al. develop an adaptive Control Lyapunov Functions and Model Predictive Control (CLF-MPC) framework in ANYmal that ensures stability and allows for online convergence of the robot's body inertial and mass parameters \cite{Minniti_12}. Sombolestan et al. present an adaptive force control for quadrupedal locomotion under uncertainties\cite{Sombolestan_12}. In addition, in \cite{Pandala_13}, a robust model predictive
control (RMPC) utilizes Reinforcement Learning (RL) to train a neural network that
outputs uncertainties for a reduced-order model. Similarly, in
the work by \cite{Sun_14}, an online training residual model is used to
correct the nominal model in MPC, which enables the Unitree A1
robot to carry a 10 kg payload.

\paragraph{Physics-Data Hybrid Control} Due to the complexity of a real-world robotic system, some parts of it cannot be expressed by functions of physical laws. To solve this problem, some researchers have used data-driven models that describe these parts of the system and combined them with physics-based models to achieve better control of real-world robots \cite{Yang_14} -\cite{Torrente_14}. 

\subsection{Contribution} 
The summary of our main contributions is as follows: 
\begin{itemize}
\item {We use a data-driven method for quadrupedal locomotion by time-series data to tackle model uncertainties. Due to the periodic locomotion characteristic, the method exhibits a remarkable capability to provide highly accurate predictions for future data.}
\item {We propose a novel control framework for quadruped robots where the data-driven error model is combined with MPC. It can help the controller to have a better estimate of future robot states even under MPC with a defect model, then adjust desired inputs for the robot.}
\item {We validate our controller in various simulations and real-world experiments on our quadruped robot Sirius-belt. The hardware tests showcase that the robot with the proposed control framework is able to trot forward stably while carrying a 20 kg un-modeled payload (84\% of body weight) as shown in Fig. \ref{forward_trotting}. Our method can not only reduce tracking errors but also the vibration amplitude of robot states under dynamic uncertainties.}
\end{itemize}

\section{MPC with Error Model} \label{MPC_Error_model}

%In this section, we present the integration of the error model, derived from the ARMAV model, into the traditional MPC framework for quadruped robots. The primary sensors commonly employed in quadruped robots are the inertial measurement unit (IMU) and encoders, which provide measurements of essential robot states, such as roll, pitch, yaw (RPY), and body height. These sensor measurements are reasonable approximations of the true values of these states. However, the estimation of the robot body displacement in the X-axis and Y-axis involves integration of sensor data, which can lead to deviations from the actual values. As a result, these integrated values are not suitable for directly fitting the error model. Consequently, our focus lies in constructing the error model based on the RPY and body height measurements, which are more reliable and less prone to integration-related errors.

In this section, we present the integration of the error model, derived from the Autoregressive Moving Average Vector (ARMAV) model \cite{Pandit_15}, into the traditional MPC framework for quadruped robots.

\begin{center}
\vspace{-0.3cm}
\begin{figure*}[htb]
\centering
\includegraphics[width=6.6in]{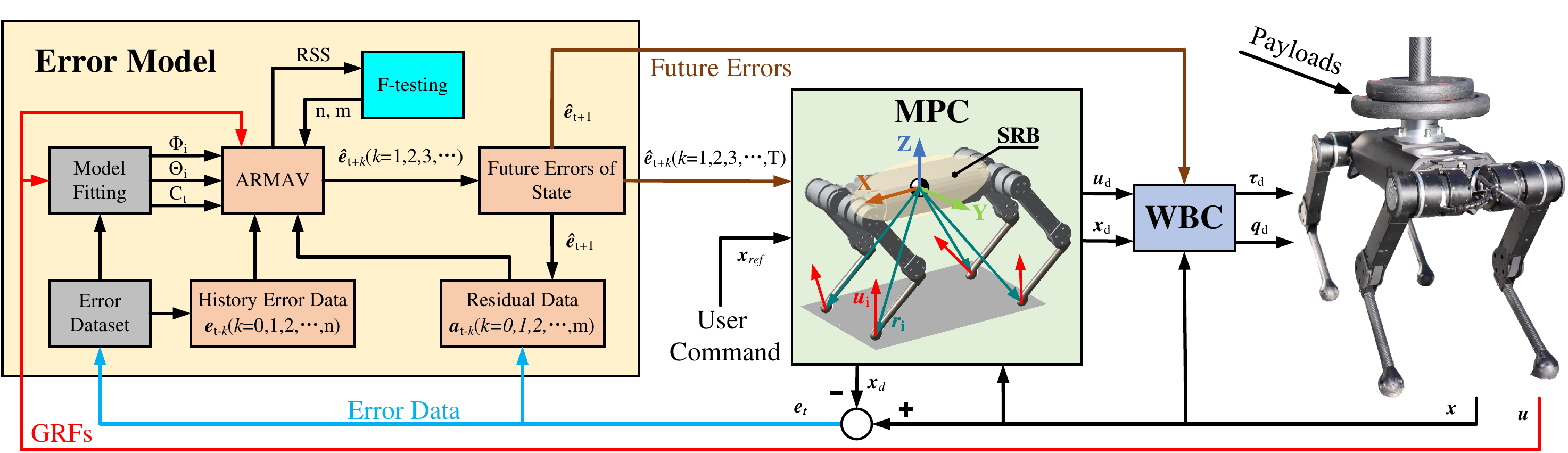}
\caption{Overview of proposed control framework. The state error data from the robot are collected to construct a data-driven error model including modeling (Section \ref{modeling}) shown in \textcolor[rgb]{0.74902,0.74902,0.74902}{grey}, determining model order (Section \ref{checking}) shown in \textcolor[rgb]{0,1,1}{blue} and making prediction (Section \ref{Model_structure}) shown in \textcolor[rgb]{0.9725,0.796,0.6784}{orange}. Then the  Predicted values of future robot state from MPC based on the reduced-order model (ignoring dynamic of legs) are adjusted by muti-steps ahead prediction of error model then outputs desired robot states ${\bm x_d}$ and GRFs (Ground Reaction Forces) ${\bm u_d}$ by optimization for WBC (Whole-body Control). WBC is also adjusted by one-step ahead prediction from the error model and uses the full-order model to calculate joint torque ${\bm \tau _d}$ and joint position ${\bm q_d}$ for a quadruped robot.}.
\label{control_framework}
\end{figure*}
\vspace{-0.5cm}
\end{center}

\subsection{Linearized Dynamic Model}
In the context of Model Predictive Control (MPC) applied to quadruped robots, commonly the robot's body is treated as a Single Rigid Body (SRB) and the state of the robot can be defined as follows:
\begin{equation}
\begin{aligned}
{\bm x}=[{\bm \theta}^T ~ {\bm p}^T ~ {\bm \omega}^T ~ {\bm v}^T ~ g]^T
\end{aligned}
\label{eq:state}
\end{equation}
Where ${\bm \theta}, {\bm p}, {\bm \omega}, {\bm v} \in \mathbb{R}^{3} $ are the Euler angles, positions, angle velocities, and linear velocities of robot body in the world frame, ${g}$ is the gravitational acceleration.

In each control cycle of the MPC algorithm, the future behavior of the robot is predicted using the floating base model, and its discrete-time formulation \cite{Di_01} is as follows:
\begin{equation}
\begin{aligned}
{\bm x_{t+1}}={A_t \bm x_t}+{ B_t \bm u_t}
\end{aligned}
\label{eq:discrete_state}
\end{equation}
Where ${\bm x_t} \in  \mathbb{R}^{13}$ is the states of the robot at time step ${t}$, ${\bm u_t} \in  \mathbb{R}^{12}$ are Ground React Forces (GRFs) which are inputs of the system at time step ${t}$, ${ A_t} \in  \mathbb{R}^{13 \times 13}$ is the matrix related with states at time step ${t}$, and ${ B_t \in  \mathbb{R}^{13 \times 12}}$ is the matrix about inputs at time step ${t}$.

The equation (\ref{eq:discrete_state}) is based on the model of the robot body after being linearized and simplified (ignoring the dynamic of legs). In real-world scenarios, practical robots inevitably exhibit deviations from this model as mentioned before. 

\subsection{Integrating Error Model into MPC}
The primary sensors commonly employed in quadruped robots are the inertial measurement unit (IMU) and encoders, which provide measurements of essential robot states, such as Euler angles and body height. These sensor measurements are reasonable approximations of the true values of these states.  Consequently, our focus lies in constructing the error model based on these sensor data.

%However, the estimation of the robot body displacement in the X-axis and Y-axis involves integration of sensor data, which can lead to deviations from the actual values. As a result, these integrated values are not suitable for directly fitting the error model.

In this paper, An ARMAV error model fitted from sensor data is implemented in traditional MPC to compensate for errors of dynamics uncertainties. By using estimated GRFs as indicators to predict potential state errors, we modify the method to improve its performance in quadruped robot applications. The modified ARMAV($n,m$) error model has the following form (derived from the procedure outlined in Section \ref{modeling} and \ref{checking}):
\begin{equation}
\begin{aligned}
\begin{split}
&{\bm e_t}-{ \Phi_1}{\bm e_{t-1}}-{ \Phi_2}{\bm e_{t-2}}+...+{ \Phi_n}{\bm e_{t-n}}=\\
&{\bm a_t}-{ \Theta_1}{\bm a_{t-1}}-{ \Theta_2}{\bm a_{t-2}}...-{ \Theta_m}{\bm a_{t-m}}+{ C_t}{\bm u_t}
\end{split}
\end{aligned}
\label{eq:error_model}
\end{equation}
Where ${\bm e_t} \in  \mathbb{R}^{4}$ is the error of robot state at time step ${t}$, ${n}$ is autoregression (AR) parameters of the model, ${m}$ is the moving average (MA) parameters of the model, $ { \Phi_i}$, $ { \Theta_i} \in  \mathbb{R}^{4 \times 4}$ are the matrix about AR and MA model respectively, ${C_t} \in  \mathbb{R}^{4 \times 12}$ is the matrix related with inputs in the error model.

Based on the error model (\ref{eq:error_model}) and obtaining predicted values of future state errors which is illustrated in Section \ref{Model_structure}, the robot states after compensation are expressed as follows:
\begin{equation}
\begin{aligned}
{\bm {\tilde x_{t+1}}}={\bm x_{t+1}}+{S}{\bm {\hat e_{t+1}}}
\end{aligned}
\end{equation}
Where ${S} \in  \mathbb{R}^{13 \times 4}$ is the select matrix for error model.

Quadruped robot locomotion entails the critical task of regulating the position and orientation of the robot's body, commonly referred to as robot states, in accordance with the desired commands. Based on the future states of the robot, MPC outputs the suitable GRFs to attain the desired robot states via optimization. This quadratic programming problem can be formulated as below:
\begin{subequations}
\begin{align}
\min_{\bm u} \quad \quad \quad &\sum _{k=0} ^{N-1} l({\bm {\tilde x_{t+k}}},{\bm { u_{t+k}}})\\
{s.t.} \quad \quad \quad &{\bm {\tilde x_{t+k+1}}} =A_t {\bm x_{t+k}} + B_t {\bm u_{t+k}} + {S}{\bm {\hat e_{t+k+1}}},\\
\nonumber &k=0,1,...,N-1\\
&{\bm x_t}=\bm x(t)\\
&g_k({\bm {\tilde x_{t+k}}},{\bm { u_{t+k}}})=0, \quad k=0,1,...,N\\
&h_k({\bm {\tilde x_{t+k}}},{\bm { u_{t+k}}})\geq0, \quad k=0,1,...,N
\end{align}
\end{subequations}
Where ${T}$ is the predict horizon, $l(\cdot)$ is the stage quadratic cost function, ${\bm x_t}$ is the current state at time step $t$, ${\bm x_{t+k}}$ is the predicted state at time step $t+k$ based on linear body dynamic, $g_k(\cdot)$ and $h_k(\cdot)$ are general equality, inequality constraints respectively.

%%\begin{center}
%\vspace{-0.3cm}
%\begin{figure}[htb]
%\centering
%\includegraphics[width=3.2in]{control framework.jpg}
%\caption{The control frameworks of proposed %controller}\label{robot_coordinate}
%\end{figure}
%\vspace{-0.5cm}
%\end{center}

The overall control framework is illustrated in Fig.~\ref{control_framework}. The procedure is that the error model is derived from the error data collected in the sensors on the robot within every control cycle and the model uses historical data to predict future state errors. Subsequently, the Controller leverages these prognostic state errors to effectuate compensatory adjustments, countering discrepancies arising from the control model. The compensation strategy is twofold: 1. The error model employs a multi-step prediction approach to mitigate disparities within the MPC; 2. One step-ahead prediction is used to compensate for the Whole-body Control (WBC).

\section{ARMAV Model}
\label{ARMAV_Model}
The objective of this section is to introduce the methodology of the ARMAV model based on time-series data.
%In traditional MPC framework of quadruped robot, the locomotion performance largely depends on ideal model. While, because of inaccuracy model parameters of robot (mass, inertia, friction, etc.), errors of model after linearization and simplified, and the errors of motor model (output torque error), the performance of robots in real world is not as good as expected. Thus, in this part, we want to introduce a data-driven error model to make a compensation of the gap.

\subsection{Model Structure} \label{Model_structure}
The ARMAV model\cite{Pandit_15} is a widely used statistical method for analyzing and predicting time series data. It has been successfully applied in various fields such as marketing prediction\cite{market_24}, manufacturing \cite{manufacturing_25}, resources management \cite{resource_26,resource_27}, and so forth. One of its key strengths lies in its ability to capture and predict periodic behavior. Given that the locomotion of our quadruped robots exhibits periodic patterns, the ARMAV model is well-suited for error prediction in this context. The formulation of the ARMAV($n,m$) model is presented below:
\begin{equation}
\begin{aligned}
{\bm z_t}-{\Phi_1}{\bm z_{t-1}}-{ \Phi_2}{\bm z_{t-2}}-...-{ \Phi_n} {\bm z_{t-n}}=
\\{\bm a_t}-{ \Theta_1}{\bm a_{t-1}}-{ \Theta_2}{\bm a_{t-2}}-...-{\Theta_m}{\bm a_{t-m}}
\end{aligned}
\label{eq:armav}
\end{equation}
where ${\bm z_t} \in  \mathbb{R}^{r}$ represents the data sample at time step ${t}$, ${\bm a_t} \in  \mathbb{R}^{r}$ is the residual modeling noise at time step ${t}$, ${ \Phi_i} \in  \mathbb{R}^{r \times r}$ is the parameters matrix related with AR model and ${ \Theta_i} \in  \mathbb{R}^{r \times r}$ is the parameters matrix about MA model.

The update of ${\bm a_t}$ and one-step ahead prediction of ARMAV($n,m$) model can be obtained as the following equation:
\begin{equation}
\begin{aligned}
{\bm {\hat{z}_{t+1}}}={ E}[{\bm z_t}|t]={\Phi_1}{\bm z_t}+{ \Phi_2}{\bm z_{t-1}}+...+{ \Phi_n}{\bm z_{t-n+1}}
\\-{ \Theta_1}{\bm a_t}-{ \Theta_2}{\bm a_{t-1}}-...-{ \Theta_m}{\bm a_{t-m+1}}
\label{eq:xt_predict}
\end{aligned}
% \end{eqnarray}
\end{equation}
\begin{equation}
\begin{aligned}
{\bm a_{t+1}}={\bm z_{t+1}}-{\bm {\hat{z}_{t+1}}}
\end{aligned}
% \end{eqnarray}
\end{equation}
Where ${\bm {\hat{z}_{t+1}}}$ is the predicted value for time step ${t+1}$.

As for k-step ahead prediction, it is derived from iteration which means that the next predicted values are based on the before predicted values as well as real data:
\begin{equation}
\begin{aligned}
{\bm {\hat{z}_{t+k}}}&={ \Phi_1}{\bm {\hat{z}_{t+k-1}}}+{ \Phi_2}{\bm {\hat{z}_{t+k-2}}}+...+{ \Phi_n}{\bm z_{t+k-n}}
\\&-{ \Theta_1}{\bm {\hat{a}_{t+k-1}}}-{ \Theta_2}{\bm {\hat{a}_{t+k-2}}}-...-{ \Theta_m}{\bm a_{t+k-m}}
\end{aligned}
\end{equation}

\begin{equation}
{\bm {\hat{a}_{t+l-m}}}=0,l=1,2,3,...,k
\label{eq:at_predict}
\end{equation}

\subsection{Modeling Method}\label{modeling}
Due to the nonlinearity of the ARMAV($n,m$) model, parameter estimation is a non-linear problem. A viable solution is to obtain the initial guess values through the inverse function of ARMAV($n,m$):
\begin{equation}
{\bm a_t}={\bm z_t}-{ I_1}{\bm z_{t-1}}-{ I_2}{\bm z_{t-2}}-...-{ I_l}{\bm z_{t-l}}...
\label{eq:inverse_function}
\end{equation}
Where ${ I_l}$ is the inverse coefficient.

In the inverse function form of the ARMAV model, the relations in ${ \Phi_i}$ and ${ \Theta_i}$ expressed by the inverse coefficients ${ I_j}$ are linear for an arbitrary ARMAV($n,m$) model. Since coefficients ${I_j}$ are AR parameters of the infinite expansion of the ARMAV model, it can be estimated well by the Least-Squares (LS) approach.

Substituting for ${\bm a_t}$ from equation (\ref{eq:inverse_function}) in equation (\ref{eq:armav}), the relation between ${\Phi_i}$, ${\Theta_i}$ and ${I_j}$ can be expressed as following:  
%%The final values of AR and MA parameters can be obtained by iteration based on the initial guess values. 
%%a initial guess values of AR and MA parameters can be known based on the inverse function:
\begin{equation}
\begin{split}
&({E}-{ \Phi_1}B-...{\Phi_n}B^n){\bm z_t}=\\
&({E}-{ \Theta_1}B-...-{\Theta_m}B^m)({E}-{ I_1}B-{ I_2}B^2-...){\bm z_t}
\end{split}
\end{equation}
Where ${E}$ is identity matrix and ${B}$ is back shift operator (for instance, ${B^i \bm z_t= \bm z_{t-i}}$).

Equating the coefficients of equal powers of $B$, the following is obtained:
\begin{equation}
\begin{aligned}
\begin{split}
&{\Phi_1}={\Theta_1}+{I_1}\\
&{\Phi_2}={\Theta_2}-{\Theta_1}{I_1}+{I_2}\\
&{\Phi_3}={\Theta_3}-{\Theta_1}{I_2}-{\Theta_2}{I_1}+{I_3}\\
&. \ . \ .\\
&{\Phi_j}={ \Theta_j}-{\Theta_1}{I_{j-1}}-{\Theta_2}{ I_{j-2}}-...-{\Theta_{j-1}}{I_1}+{I_j}
\end{split}
\label{eq:get_phi}
\end{aligned}
\end{equation}
Where it is assumed that ${ \Theta_j}=0$ for ${j>m}$ and ${\Phi_j}=0$ for ${j>n}$.

According to equation (\ref{eq:get_phi}), for ${j> \max (n,m)}$ the particular case of it can be derived:
\begin{equation}
({E}-{\Theta_1}B-{\Theta_2}B^2-...-{\Theta_m}B^m){I_j}=0
\label{eq:get_theta}
\end{equation}

It is straightforward that once estimated values of inverse coefficients are known the initial values of AR parameters as well as MA parameters can be derived from equation (\ref{eq:get_phi}) and (\ref{eq:get_theta}). The initial guess of ${I_j}$ is given from pure AR($p$) model:
\begin{equation}
{\bm z_t}={\Phi_1}{\bm z_{t-1}}+{ \Phi_2}{\bm z_{t-2}}+...+{\Phi_p}{\bm z_{t-p}}+{\bm a_t}
\end{equation}

Based on the estimates of ${ I_j}$ in AR($p$) model, ${ \Theta_i}$ is derived from equation (\ref{eq:get_theta}). Then by substituting these ${ \Theta_i}$ in equation (\ref{eq:get_phi}), the explicit solution of ${\Phi_i}$ is got. Since the initial values of ${\Theta_i}$ are derived from equation (\ref{eq:get_theta}) that requires ${j> \max (n,m)}$, for getting all the ${ \Theta_i}$, it is reasonable to take the order  of AR($p$) model as:
\begin{equation}
p=\max (n,m) + m
\end{equation}

\subsection{Checking Criterion}\label{checking}
One of the reasons why the ARMAV model is so popular is that it is always possible to represent the dynamics of a stable stationary stochastic system with the ARMAV($n, m$) model, which means that only if the order of system $n$ and $m$ increases can the error of the model decrease into as small as we want. This flexibility allows for fitting the data to different orders of the ARMAV model based on the complexity of the system and the desired level of accuracy.

% \begin{algorithm}[t]
% \SetAlgoLined
% \KwIn{input parameters A, B, C}%输入参数
% \KwOut{output result}%输出
% \KwResult{Write here the result }
%  initialization\;
%  \While{While condition}{
%   instructions\;
%   \eIf{condition}{
%    instructions1\\
%    instructions2\\
%    }{
%    instructions3\;
%   }
%  }
%  \caption{How to write algorithms}
% \end{algorithm}

% \end{algorithmic}

% \begin{center}
% \vspace{-0.5cm}
% \begin{figure}[htb]
% \centering
% \includegraphics[width=3.3in]{F testing.jpg}
% \caption{Procedure diagram of F-testing to determine the order of ARMAV model}\label{F_testing}
% \end{figure}
% \vspace{-0.4cm}
% \end{center}

The reduced trend of residual sum of squares (RSS) is a good indicator to evaluate the adequacy of the model, when RSS drops significantly which means that the model may not be enough and the order of ARMAV($n, m$) should be increased. F checking criterion \cite{Pandit_15} is an indicator to evaluate the drop trend of RSS :
\begin{equation}
\begin{aligned}
F=\frac{A_1-A_0}{s} \div \frac{A_0}{N-r} \sim F(s,N-r)
\end{aligned}
\label{eq:f_testing}
\end{equation}
Where ${A_0}$ is the smaller RSS in the unrestricted model, ${A_1}$ is the larger RSS in the restricted model, ${s}$ is the number of restricted parameters, ${N}$ is the number of samples, ${r}$ is the number of estimated parameters, and ${F(s,N-r)}$ is F-distribution with ${s}$ and ${N-r}$ degrees of freedom. If ${F<F_{s,N-r}^{\alpha}}$, RSS can be regarded as drop significantly with confidence ${\alpha}$.

The procedure of the F-testing method based on equation (\ref{eq:f_testing}) is used to choose the order of the ARMAV model. The first step is to determine the AR order $n$ by comparing the RSS of ARMAV($2k+2, 2k+1$) and ARMAV($2k,2k-1$) starting from $k=1$. Then, the MA order $m$ is reduced to determine the desired value by a similar method. At last, the autocorrelation of residuals with lag $l$ should be confirmed:
\begin{equation}
\begin{aligned}
\rho_l=\frac{\sum_{t=1}^{N-l} a_ta_{t+l}}{N-l} \div \frac{\sum_{t=1}^{N} a_ta_t}{N}
\end{aligned}
\label{eq:autocorr}
\end{equation}

\section{Simulation}\label{section:simulation}
This section focuses on evaluating the performance of the proposed controller in an open-source simulation environment \cite{Di_17}.

\subsection{Inaccuracy Model Parameters Situation}\label{simulation_1}
In this case, to assess the performance of the proposed controller when model parameters are inaccurate, two groups have been established: 1. the control model is perfect (ground truth case); 2. the mass of the model is set to 34.7 kg (the real mass of the robot is 23.7 kg). 

\iffalse
\begin{figure}[htb]
  \centering
\vspace{-0.3cm}
 \subfigure[Ground truth case]{
    \label{CaRe1} %% label for second subfigure
    \includegraphics[width=0.23\textwidth]{perfect_belt_armav_body_height.pdf}}
  \subfigure[Inaccurate mass case]{
    \label{CaPe2} %% label for first subfigure
    \includegraphics[width=0.23\textwidth]{m34_belt_armav_body_height_2.pdf}}
    
\vspace{-3mm}
  \caption{The comparison between actual error of body height and predicted value from error model.}
  \label{sim_bodyheight_predict} %% label for entire figure
\vspace{-0.2cm}
\end{figure}
%%robot state
\fi

\begin{figure}[htb]
  \centering
\vspace{-0.1cm}
 \subfigure[CoM height of ground truth case]{
    \label{CaRe1} %% label for second subfigure
    \includegraphics[width=0.44\textwidth]{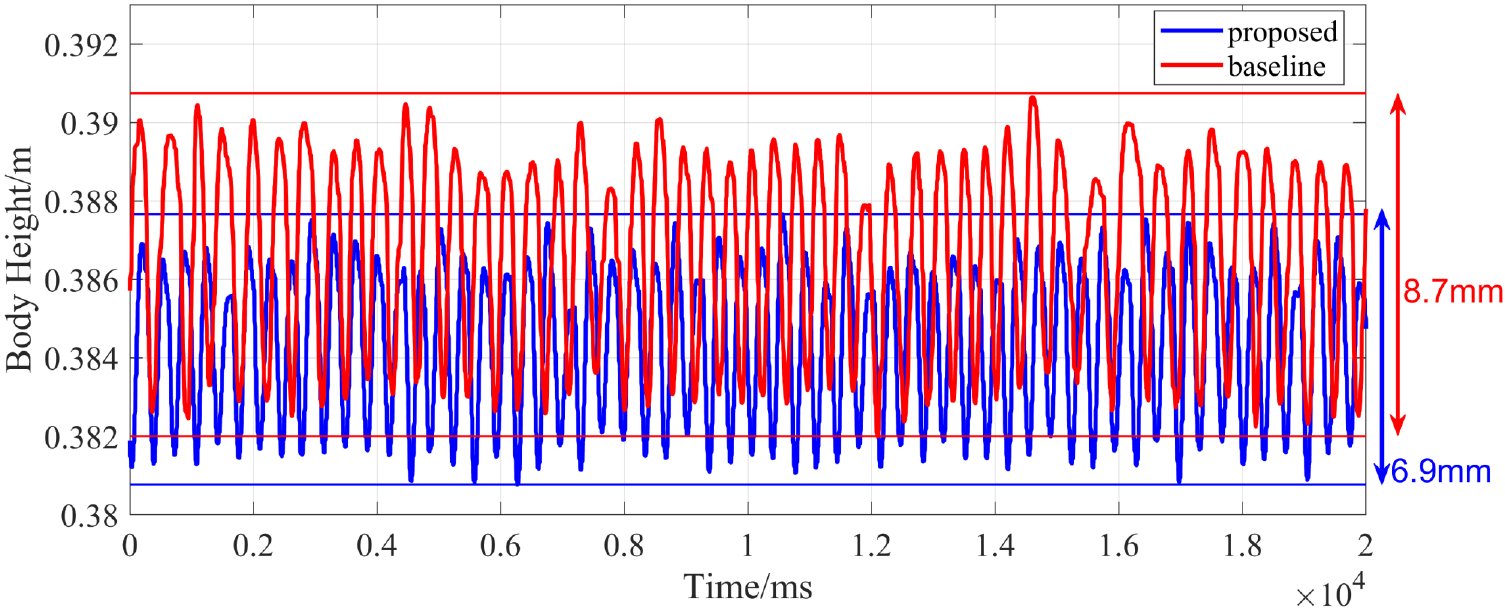}}
  \subfigure[CoM height of inaccurate mass case]{
    \label{CaPe2} %% label for first subfigure
    \includegraphics[width=0.44\textwidth]{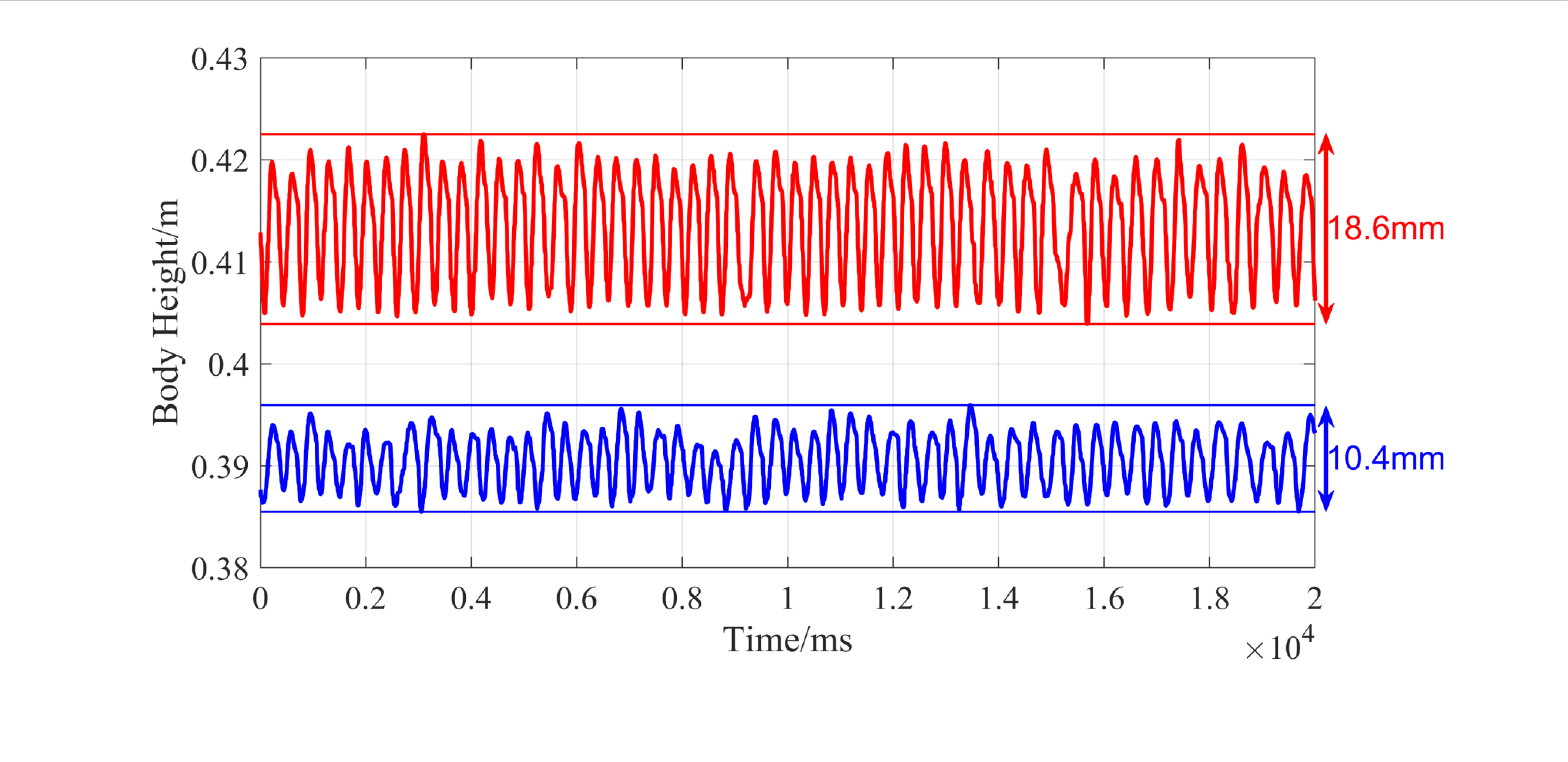}}
\vspace{-3mm}
  \caption{The vibration of body height during trotting in place. In these two cases, the desired body height is 0.38 m.}
  \label{sim_inaccurate_parameter_performance} %% label for entire figure
\vspace{-0.3cm}
\end{figure}

\begin{figure}[htb]
  \centering
\vspace{-0.2cm}
 \subfigure[Ground truth case]{
    \label{CaRe1} %% label for second subfigure
    \includegraphics[width=0.23\textwidth]{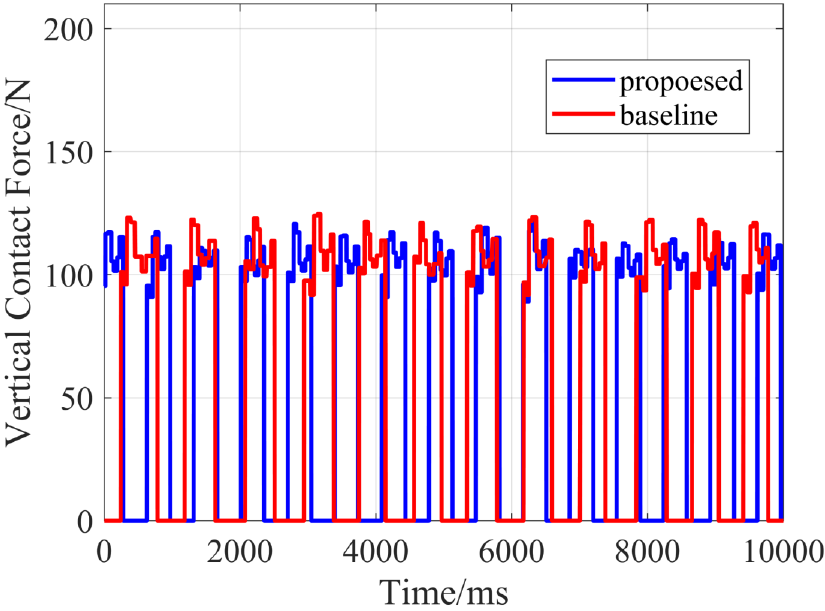}}
  \subfigure[Inaccurate mass case]{
    \label{CaPe2} %% label for first subfigure
    \includegraphics[width=0.23\textwidth]{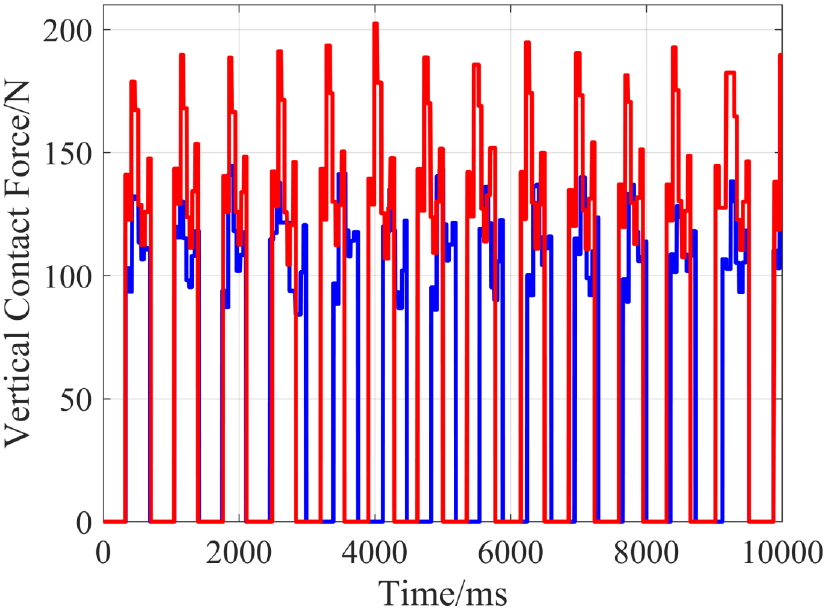}}
\vspace{-3mm}
  \caption{GRFs (Ground Reaction Forces) of one leg in the vertical direction in simulation. The data are from the left front leg of the robot and the other legs have a similar trend. In the case of inaccurate mass, it is obvious that baseline MPC generates larger GRFs (red line) due to the defect model while the propose controller has a great effect in adjusting it (blue line).}
  \label{sim_GRF} %% label for entire figure
\vspace{-0.3cm}
\end{figure}

The locomotion performances of quadruped robots in two cases are compared based on the displacement of CoM and the GRFs in the vertical direction. The comparative outcomes are visually presented in Fig.~\ref{sim_inaccurate_parameter_performance} and Fig.~\ref{sim_GRF}.

In the simulation, the robot's desired body height is set to 0.38 m. Fig.~\ref{sim_inaccurate_parameter_performance} shows that the error model helps to bring the robot's state closer to the desired value, and reduces the vibration amplitude of the body height for both cases. For the perfect model case, errors mostly arise from the model's simplified and linearized process. Therefore, the error model only achieves a slight improvement of locomotion by reducing 21.7\% vibration of CoM, from 8.7 mm to 6.9 mm. In the inaccurate mass case, there is an 11 kg gap in the mass parameter between the robot and the model. This is the reason why the body height vibrates largely. By adding the error model, this model inaccuracy is compensated and its CoM vertical vibration is decreased by 41.4\%, from 18.6 mm to 10.9 mm. Fig.~\ref{sim_GRF} shows that the error model helps MPC adjust the GRFs to achieve more suitable values for such improvement. In conclusion, the error model can improve the locomotion performance of the robot in both cases, and the improvement is more obvious under the inaccurate model parameter situation.

\subsection{Un-modeled Payload Situation}
For the quadruped robot in the real world, transporting various payloads is one of its common applications. However, the parameters of payload such as mass are random depending on application scenarios and it is difficult to be included in the control model in advance. To simulate such a situation, in this part, an un-modeled payload is applied to the robot body.

\begin{figure}[htb]
  \centering
\vspace{-0.2cm}
 \subfigure[The vertical displacement of body COM]{
    \label{sim_payload_bodyheight} %% label for second subfigure
    \includegraphics[width=0.45\textwidth]{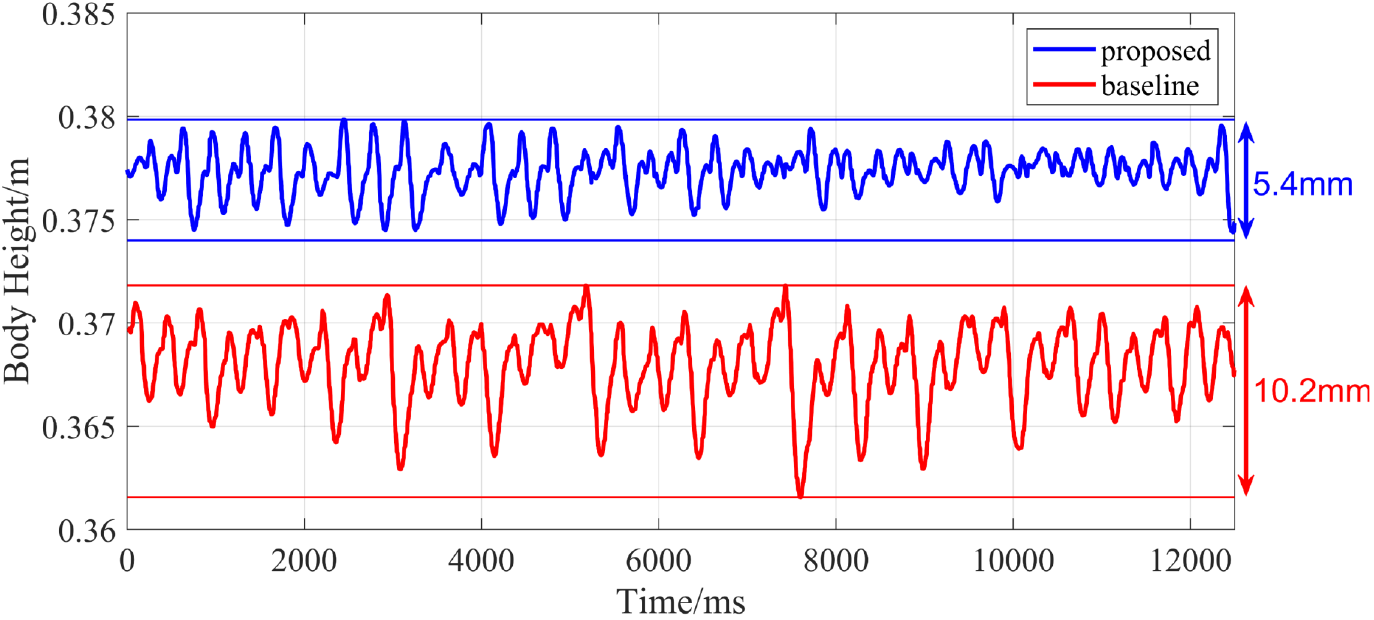}}
  \subfigure[Baseline (fall down)]{
    \label{sim_baseline} %% label for first subfigure
    \includegraphics[width=0.184\textwidth]{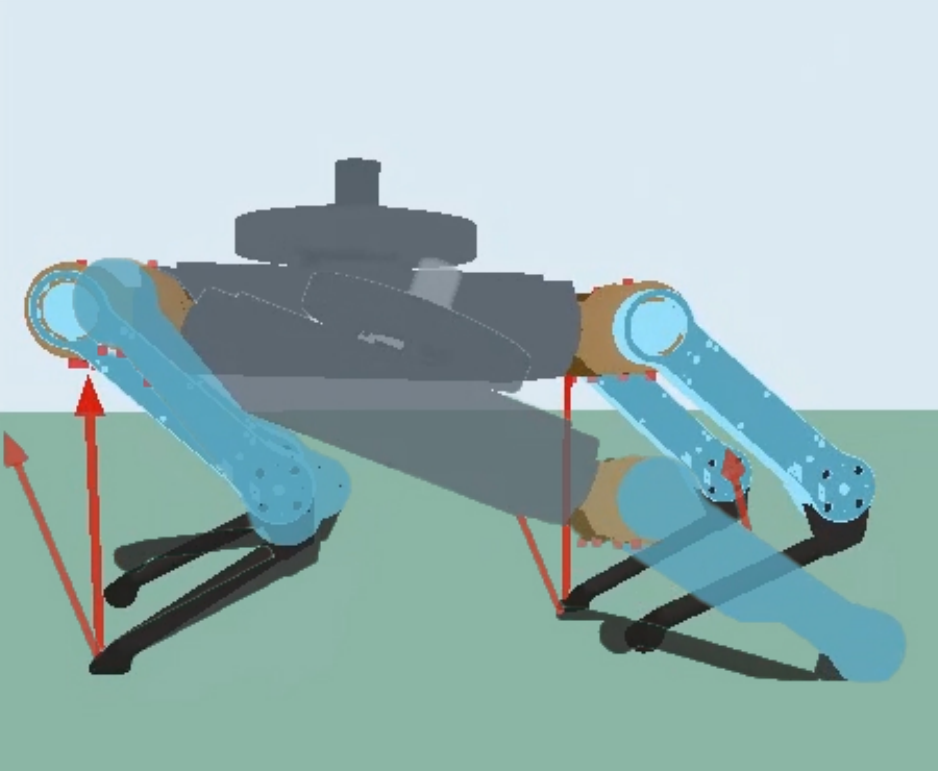}}
  % \subfigure[Baseline(fall down)]{
  %   \label{sim_baseline_fall} %% label for first subfigure
  %   \includegraphics[width=0.15\textwidth]{baseline_payload8_falldown.png}}
  \subfigure[Proposed controller]{
    \label{sim_proposed} %% label for first subfigure
    \includegraphics[width=0.18\textwidth]{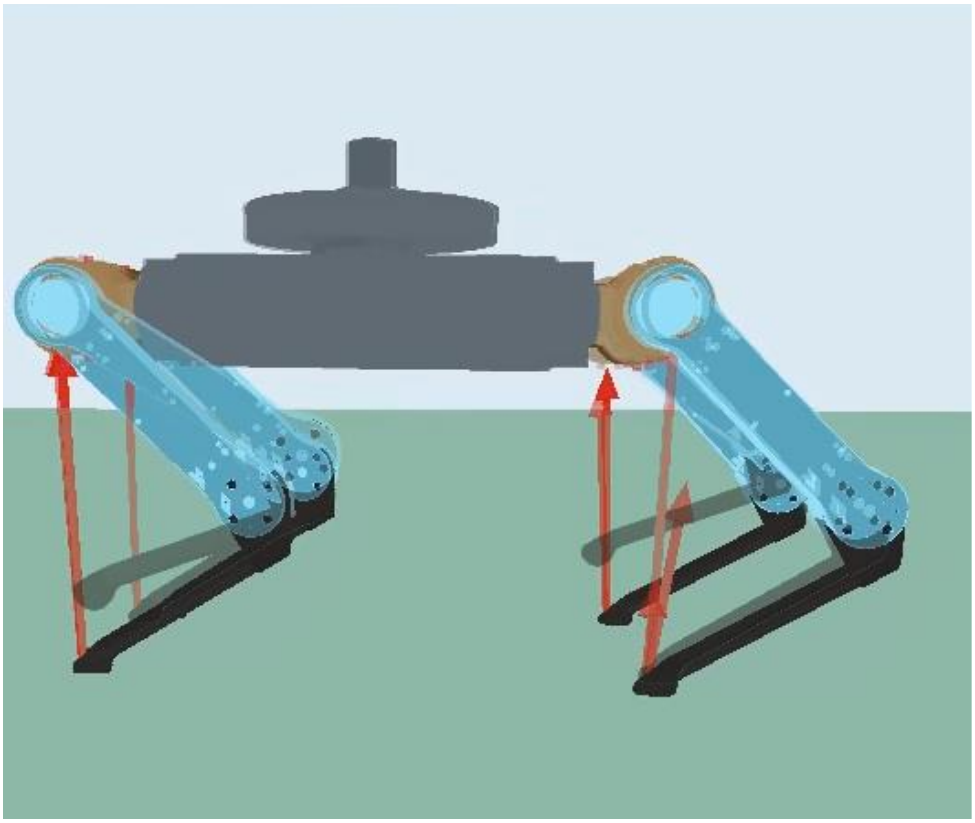}}
\vspace{-3mm}
  \caption{The simulation results of the robot with an un-modeled 8 kg payload. In this simulation, the desired height of CoM is also 0.38 m. }
  \label{2d_planar_model} %% label for entire figure
\vspace{-0.3cm}
\end{figure}

\begin{center}
\vspace{-0.3cm}
\begin{figure*}[htb]
\centering
\includegraphics[width=5.8in]{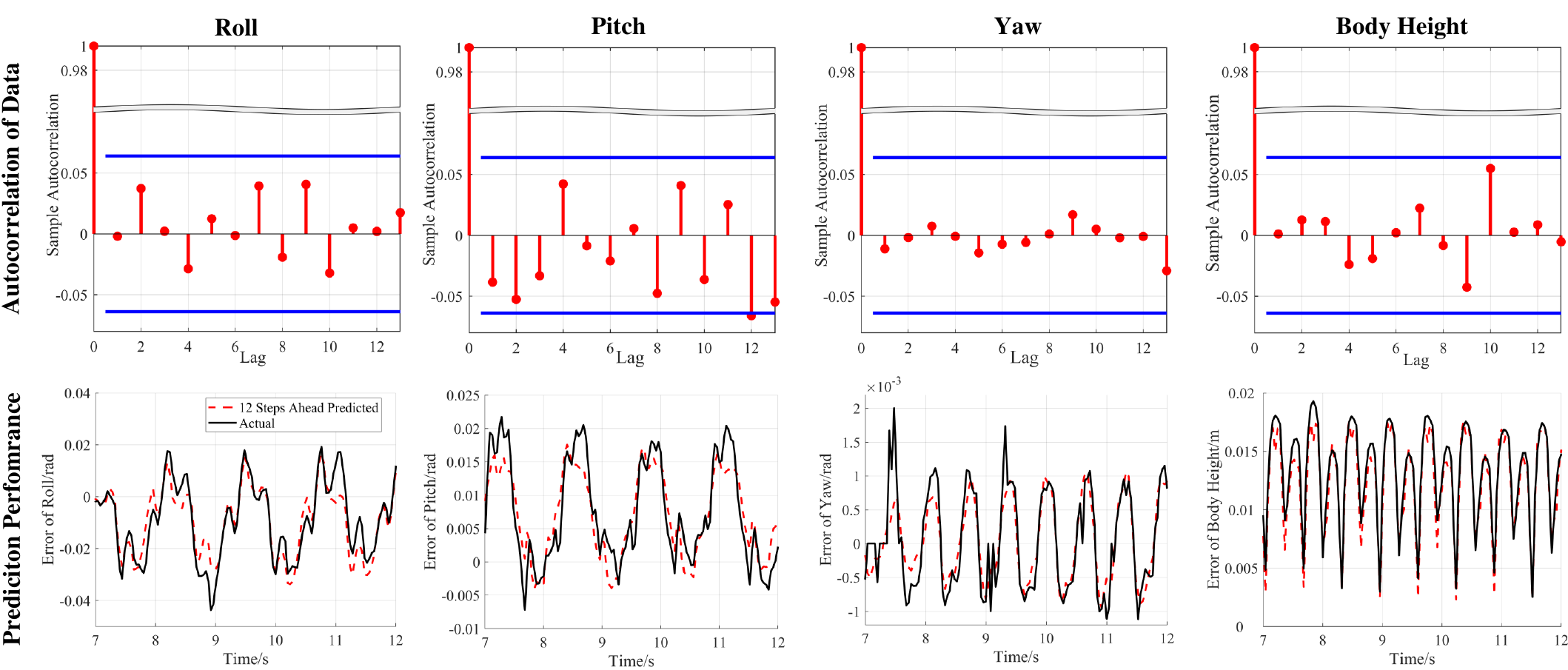}
\caption{The autocorrelation and 12 (which is also the MPC predict horizon) steps ahead prediction performance of the error model in the real robot. The data are collected when the quadruped robot is trotting in place with a 10 kg payload. The figure indicates that the error model is sufficient for data and predicts future state errors accurately in Euler angles (Roll, Pitch Yaw) and body height.}
\label{real_error_model}
\end{figure*}
\vspace{-0.5cm}
\end{center}

In the simulation, an 8 kg payload is exerted on the robot body, which is excluded from the control model. The vibration of body height under this situation is shown in Fig.~\ref{sim_payload_bodyheight}. The baseline controller fails to restore the robot to the desired state, as the payload is not modeled, and the body height remains below the desired value of 0.38 m. On the other hand, the proposed controller has a noticeable effect in raising the body height to the desired value while simultaneously reducing the vibration by 47.1\%. Additionally, as shown in Fig.~\ref{sim_baseline} and Fig.~\ref{sim_proposed}, the performance of the robot with the baseline controller is less stable compared to the robot with the proposed controller and it falls down after 30 seconds of trotting. In conclusion, the proposed controller can make the robot attain the desired body height, reduce the vibration of CoM, and ensure stable trotting even under an unknown payload in the simulation.

%\begin{center}
%\vspace{-0.5cm}
%\begin{figure}[htb]
%\centering
%\includegraphics[width=3.4in]{sim_payload3.3_body_height.jpg}
%\caption{The vertical displacement of body COM under 8 kg payload in simulation}\label{robot_coordinate}
%\end{figure}
%\vspace{-0.7cm}
%\end{center}

\section{Experimental Results}
\label{Experiment}
In this section, we validate our proposed controller in real-world experiments by comparing it with linear MPC.

\subsection{Experimental Setup}
The experiments are all implemented in our quadruped robot Sirius-Belt. Each leg of the robot consists of three degrees of freedom (DoFs) actuated by three electric motors. So Sirius-Belt has a total of 18 DoFs. The dimension (Length$\times$Width$\times$Height) of the robot is $0.78\times0.37\times0.54$ m (fully standing). The maximum torque for knee joint actuators (Gear Ratio: 9 to 1) and hip joint actuators (Gear Ratio: 6 to 1) are 38 Nm and 30 Nm respectively. The controller runs in an onboard Upboard (ATOM x5-Z8350). 

\subsection{Error Model Setup}
Real-world data are collected from the actual robotic system during trotting. By equation (\ref{eq:autocorr}), the residuals autocorrelation of Euler angles and body height in the error model is depicted in the up part of Fig.~\ref{real_error_model}. It can be seen that the autocorrelation of residuals between different times are all within the boundary of ${ -\frac {2}{\sqrt{N}}}$ and ${ \frac {2}{\sqrt{N}}}$ (blue lines in Fig.~\ref{real_error_model}), which can prove that residuals in error model can be regarded as uncorrelated data with mean zero and variance ${N^{-1}}$.
The below part of Fig.~\ref{real_error_model} shows 12 (which is our MPC predicted horizon) steps ahead of predicted values from the error model in Euler angles and body height which are very close to the true values of the real-world robot.
%\begin{center}
%\vspace{-0.3cm}
%\begin{figure*}[htb]
%\centering
%\vspace{-0.2cm}
% \subfigure[]{
%    \label{CaRe1} %% label for second subfigure
%    \includegraphics[width=0.231\textwidth]{autocorr_roll.jpg}}
% \subfigure[]{
%    \label{CaPe2} %% label for first subfigure
%    \includegraphics[width=0.231\textwidth]{autocorr_pitch.jpg}}
% \subfigure[]{
%    \label{CaRe3} %% label for second subfigure
%    \includegraphics[width=0.231\textwidth]{autocorr_yaw.jpg}}
% \subfigure[]{
%    \label{CaPe4} %% label for first subfigure
%    \includegraphics[width=0.231\textwidth]{autocorr_height.jpg}}
% \subfigure[]{
%    \label{CaRe1} %% label for second subfigure
%    \includegraphics[width=0.231\textwidth]{real_roll_error.jpg}}
% \subfigure[]{
%    \label{CaPe2} %% label for first subfigure
%    \includegraphics[width=0.231\textwidth]{real_pitch_error.jpg}}
% \subfigure[]{
%    \label{CaRe3} %% label for second subfigure
%    \includegraphics[width=0.231\textwidth]{real_yaw_error.jpg}}
% \subfigure[]{
%    \label{CaPe4} %% label for first subfigure
%    \includegraphics[width=0.231\textwidth]{real_height_error.jpg}}
%\vspace{-2mm}
%\centering
%\caption{Experiment Result}
%\label{control_framework}
%\end{figure*}
%\vspace{-0.5cm}
%\end{center}

\subsection{Results}
In the first hardware experiment, the payload from 5 kg to 15 kg is added on the Sirius-Belt during trotting in place one by one. Concurrently, the data of real robot states in the baseline controller and proposed controller are recorded to evaluate their performance. The body height of the robot during the experiment is shown in Fig.~\ref{experiment_result}. 

Table.~\ref{table:result} shows the performance of the proposed controller and baseline in the real-world experiment by Mean Absolute Error (MAE) and Mean Squared Error (MSE). It reveals that the proposed controller exhibits improved state tracking performance for roll, pitch, and body height in comparison to the baseline controller. However, it indicates that no significant effect is observed in the yaw state due to its pretty small errors.

\begin{table}[H]
\center
\setlength{\tabcolsep}{3mm}{
\caption{The state tracking performance MAE(MSE) of Sirius-belt carrying variable payload experiment}\label{table:result}
\begin{tabular}{c|c|c|c|c}
Method & Roll & Pitch & Yaw & Body Height \\ \hline \hline 
Baseline & 0.008 & 0.014 & 3e-4 & 5kg: 0.011(1.1e-4)\\
 & (1.6e-4) & (2.4e-4) &  & 10kg:  0.02(3.9e-4)\\
 & & & & 15kg: 0.03(0.001)\\
Proposed & \textbf{0.005} & \textbf{0.008} & 3e-4 & 5kg: \textbf{0.002(7.0e-6)}\\
 & \textbf{(4.6e-5)} & \textbf{(7.6e-5)} & & 10kg: \textbf{0.001(3.4e-6)}\\
 & & & & 15kg: \textbf{0.001(2.4e-6)} \\
\hline \hline 
\end{tabular}}
\end{table}

The second experiment is to evaluate the performance of the proposed controller in the robot's forward trotting. The results shown in Fig.~\ref{forward_trotting} prove that the proposed controller enables the Sirius-Belt carrying a 20 kg un-modeled payload to achieve a stable forward trotting, where the mean of CoM height is 0.397 m (the desired value is 0.4 m) with standard deviation (STD) of 0.0011 m. However, the counterpart with the baseline controller is unstable even in a 10 kg payload (mean = 0.369 m and STD = 0.0035 m).

The obtained results validate the effectiveness of the proposed controller in facilitating stable trotting for the robot, even when confronted with un-modeled payloads. Notably, the robot with the proposed controller consistently maintains the desired body height throughout the experiments. This capability proves advantageous when the robot is tasked with carrying heavy loads, as it prevents excessive strain on the knee actuators. 

\section{CONCLUSIONS}
\label{Conclusion}
In this paper, we present a novel controller that unifies MPC with a data-driven error model for improving the locomotion of quadruped robots. The error model is established based on the real data from sensors by the ARMAV model. This model has demonstrated an ability to accurately forecast forthcoming robot state errors by leveraging historical data and it proactively compensates for the state tracking of MPC and WBC. Besides, in the proposed controller, the error model runs concurrently with MPC, while it doesn't change control laws in MPC. Through such ways, the error model can increase the state tracking performance of MPC and will not eliminate the MPC's own robustness. The proposed controller can, to some degree, deal with the problems caused by the model's inaccuracy, linearization, and simplification in MPC. The simulation results of Section \ref{section:simulation} show that the proposed controller significantly improves the locomotion performance of the quadruped robot during trotting both in situations of inaccurate model parameters and carrying payloads. Finally, the hardware test validate that the Sirius-Belt carrying a 20 kg un-modeled payload with the proposed controller can achieve more stable forward trotting compared with the baseline controller.

\begin{center}
\vspace{-0.3cm}
\begin{figure}[htb]
\centering
\includegraphics[width=3.4in]{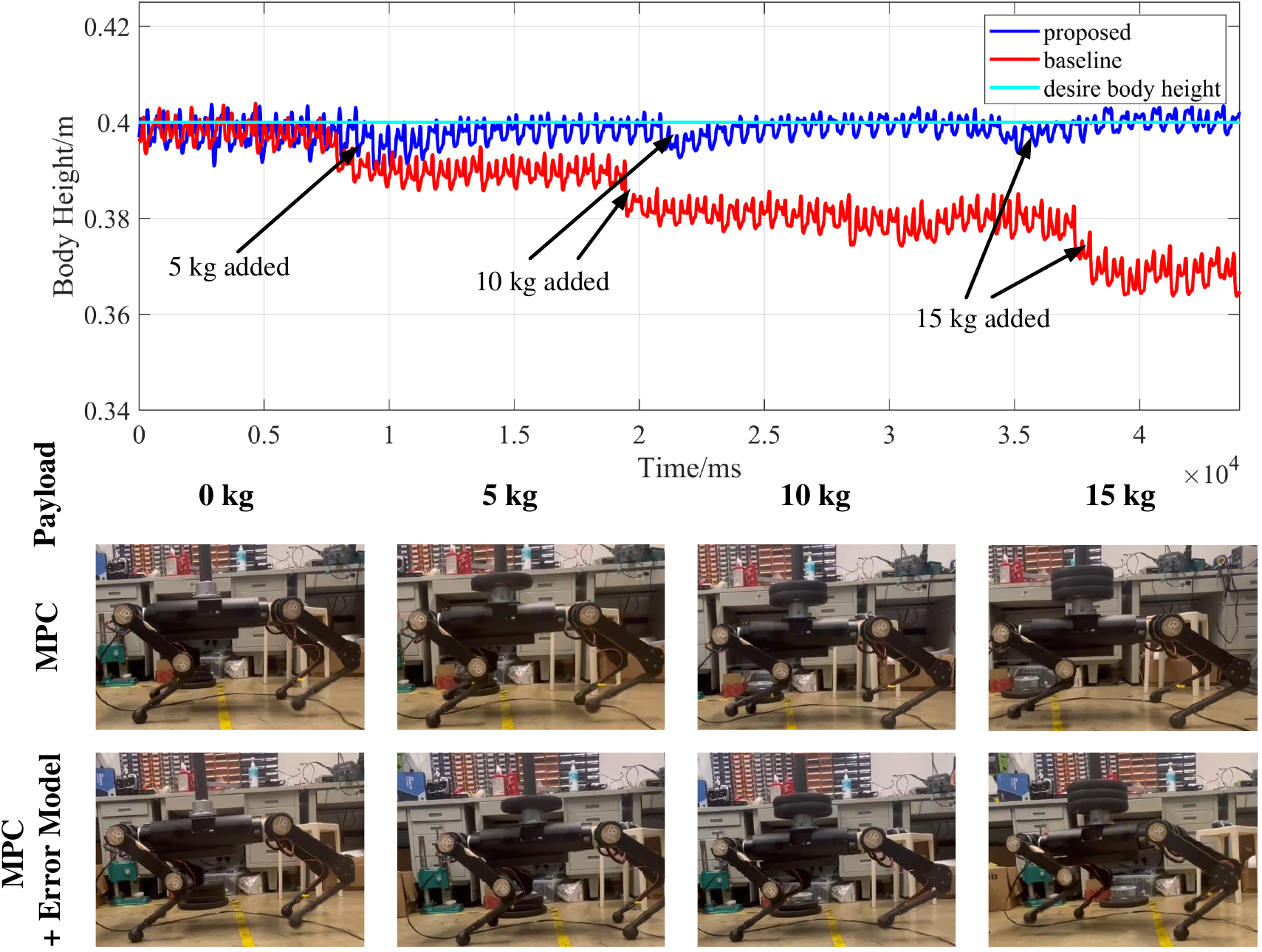}
\caption{The trotting in stance experiment result. In the experiment, the 5 kg barbell pieces are put on the robot during trotting one by one. The above part demonstrates the body height vibration versus time and the below part contains the snapshots of the quadruped robot during the test.}
\label{experiment_result}
\end{figure}
\vspace{-0.3cm}
\end{center}

%\begin{center}
%\vspace{-0.3cm}
%\begin{figure*}[htb]
%\centering
%\vspace{-0.2cm}
% \subfigure[]{
%    \label{CaRe1} %% label for second subfigure
%    \includegraphics[width=0.231\textwidth]{autocorr_roll.jpg}}
% \subfigure[]{
%    \label{CaPe2} %% label for first subfigure
%    \includegraphics[width=0.231\textwidth]{autocorr_pitch.jpg}}
% \subfigure[]{
%    \label{CaRe3} %% label for second subfigure
%    \includegraphics[width=0.231\textwidth]{autocorr_yaw.jpg}}
% \subfigure[]{
%    \label{CaPe4} %% label for first subfigure
%    \includegraphics[width=0.231\textwidth]{autocorr_height.jpg}}
% \subfigure[]{
%    \label{CaRe1} %% label for second subfigure
%    \includegraphics[width=0.231\textwidth]{real_roll_error.jpg}}
% \subfigure[]{
%    \label{CaPe2} %% label for first subfigure
%    \includegraphics[width=0.231\textwidth]{real_pitch_error.jpg}}
% \subfigure[]{
%    \label{CaRe3} %% label for second subfigure
%    \includegraphics[width=0.231\textwidth]{real_yaw_error.jpg}}
% \subfigure[]{
%    \label{CaPe4} %% label for first subfigure
%    \includegraphics[width=0.231\textwidth]{real_height_error.jpg}}
%\vspace{-2mm}
%\centering
%\caption{Experiment Result}
%\label{control_framework}
%\end{figure*}
%\vspace{-0.5cm}
%\end{center}

\addtolength{\textheight}{0cm}   % This command serves to balance the column lengths

%%%%%%%%%%%%%%%%%%%%%%%%%%%%%%%%%%%%%%%%%%%%%%%%%%%%%%%%%%%%%%%%%%%%%%%%%%%%%%%%

%%%%%%%%%%%%%%%%%%%%%%%%%%%%%%%%%%%%%%%%%%%%%%%%%%%%%%%%%%%%%%%%%%%%%%%%%%%%%%%%



% \section*{ACKNOWLEDGMENT}

% The preferred spelling of the word ÒacknowledgmentÓ in America is without an ÒeÓ after the ÒgÓ. Avoid the stilted expression, ÒOne of us (R. B. G.) thanks . . .Ó  Instead, try ÒR. B. G. thanksÓ. Put sponsor acknowledgments in the unnumbered footnote on the first page.

% %%%%%%%%%%%%%%%%%%%%%%%%%%%%%%%%%%%%%%%%%%%%%%%%%%%%%%%%%%%%%%%%%%%%%%%%%%%%%%%%

% References are important to the reader; therefore, each citation must be complete and correct. If at all possible, references should be commonly available publications.

\end{document}